\documentclass{article}
 \usepackage{graphicx}
\usepackage{amsmath, amssymb, amsthm, enumerate}
\usepackage{lscape}
\usepackage{caption}
\usepackage{subfigure}
\usepackage[utf8]{inputenc}

\usepackage{amsmath,amssymb,amsthm,enumerate,mathrsfs}

\theoremstyle{definition}

\numberwithin{equation}{section}

\usepackage[round]{natbib}

\begin{document}
\makeatletter

\begin{center}
\large{\bf A Sport Tournament Scheduling by Genetic Algorithm with Swapping Method}
\end{center}\vspace{5mm}
\begin{center}

\textsc{Tinnaluk Rutjanisarakul$^{1}$ and Thiradet Jiarasuksakun$^{2}$ \\
Department of Mathematics, Faculty of Science\\
King Mongkut's University of Technology Thonburi (KMUTT) \\
126 Pracha Uthit Rd, Bang Mod, Thung Khru, Bangkok, 10140, THAILAND
}\end{center}

\vspace{2mm}

\footnotesize{
\noindent\begin{minipage}{14cm}
{\bf Abstract:}
A sport tournament problem is considered the Traveling Tournament Problem (TTP). One interesting type is the mirrored Traveling Tournament Problem (mTTP). The objective of the problem is to minimize either the total number of traveling or the total distances of traveling or both. This research aims to find an optimized solution of the mirrored Traveling Tournament Problem with minimum total number of traveling. The solutions consisting of traveling and scheduling tables are solved by using genetic algorithm (GA) with swapping method. The number of traveling of all teams from obtained solutions are close to the lower bound theory of number of traveling. Moreover, this algorithm generates better solutions than known results for most cases.

\end{minipage}
 \\[5mm]


\hbox to14cm{\hrulefill}\par

\footnote{ 
The author would like to thank Human Resource Development in Science Project (Science Achievement Scholarship of Thailand, SAST) for financial support.
}
\footnote{
Corresponding Author
}

\section{Introduction}
\label{intro}

There are many types of graph tournaments such as round robin tournament, double round robin tournament, single-elimination tournament, double-elimination tournament, etc. The differences among these tournaments are scheduling, number of matches, and format of the competitions. For example, the FIFA world cup in the final tournament has two stages: group stage and knockout stage. In the group stage, there are eight groups and each group contains four teams. Each group plays a round robin tournament, in which each team plays one game against every other teams within the group. Then only top two teams in each group would go to the knockout stage. The knockout stage is a single-elimination tournament, in which teams play against each other in one-off matches. Another example is the tennis competition consisting of four most important annual grand slam championships. The differences among these championships are the weathers and court types but all championships are single-elimination tournament, in which the winner must win seven games consecutively. However, almost all other sport leagues, e.g. soccer, basketball, football, baseball, etc., are double round robin tournament, in which each team plays against each other team in two games: one as home game and another one as away game. 

The scheduling problem of soccer league is more popular than other sport-scheduling problem because most countries have formed their own soccer leagues. Each soccer league has usually organized a double round robin tournament. Each playing team has to find sponsors to support their costs for a tournament such as advertisement, traveling and accommodation. Therefore it would be beneficial to all if the tournament organizer could help to reduce some costs. All teams do not want to waste their time and money in transportation as a consequence of poor scheduling of the games. In a sport league, there are many variables affecting the costs of each team such as means of traveling, distances of traveling, number of traveling, sequences of traveling, etc. Thus the tournament scheduling problem affecting the traveling cost has become an important class of optimization problems for many years. 

The scheduling problems in sports are known in the literatures as traveling tournament problem proposed by \cite{en01}. Many educators were interested in scheduling problem and aimed to study the Traveling Tournament Problem differently (e.g. \cite{ah10,bi06,fa98,ru07,cz05}). They focused on improving method for solving Traveling Tournament Problem such that their results in scheduling were better approximation than preceding methods such as integer programming, canonical schedules found by Hamiltonian cycles, etc. Mostly, their objective is to minimize the total distance traveled by all teams. However, \cite{ur06} studied the mirrored traveling tournament problem and found the maximizing breaks and bounding number of traveling of solutions for each number of teams.
In this research, the objective is to minimize the total number of traveling of all teams by using Genetic Algorithm with swapping process. The swapping process was used to apply the crossover process in genetic algorithm. Moreover, the obtained solutions were compared with the known results and the bounding number of travelings of \cite{ur06}.

\section{Problem description}
\label{prob_des}

The number of traveling of each team is the total number of traveling starting from its home city and return there after the end of tournament. A game at its home city is called home game and a game at the opponent's city is called away game. Each team has at most one game in a week and each week must have equal number of home games and away games. \\

The objective is to find a tournament with minimum total number of traveling of all teams, satisfying the following constraints: 
\begin{itemize}
\item The tournament scheduling here is the mirrored Traveling Tournament Problem (mTTP).
\item The tournament problem here consists of 2 parts: team traveling and scheduling. The traveling of tournament refers to traveling of all teams. The scheduling of tournament refers to sheduling of all games of every team.
\item Each team does not go back to its home city after away game if it does not have a home game in a week after.
\item TTP's condition: The traveling sequence of each team is allowed to have at most three consecutive home games or three consecutive away games.
\end{itemize}

\section{Notations and Algorithm}
\label{Nota_Obs}
\subsection{Notations}
\label{Nota}

In this work, the tournament problem is separated into two parts: team traveling and team scheduling.\\

The team traveling and scheduling of $n$ team tournament are represented by tables $A$ and $B$, respectively, where $n$ is even. The team traveling table $A$ consists of $n$ traveling sequences of each $n$ teams. The traveling sequence of team $i$ is a sequence of 0's and 1's, where 0 stands for home game and 1 stands for away game. The element in the table $A(i,j)$ represents a type of game of team $i$ on week $w_j$. For example, one possible traveling sequence is shown in Table~\ref{t_seq}. \\
\begin{table}[h]
\centering
\caption{One possible traveling sequence of team no. $1$} \label{t_seq}
\begin{tabular}{ |c|| c c c c| c c c c| }
\hline
Team No.& $w_1$ & $w_2$ & ... & $w_{n-1}$ & $w_{n}$ & $w_{n+1}$ & ... & $w_{2n-2}$ \\ \hline \hline
 1 & 0 & 1 & ... & 0 & 1 & 0 & ... & 1 \\
\hline
\end{tabular}
\end{table} 

From the problem description, each traveling sequence may have at most three consective 0's or three consecutive 1's. After all traveling sequences are obtained, the team traveling can be formed. An example of four team tournament is shown in Table~\ref{tab_tseq}.
\clearpage
\begin{table}[h]
\centering
\caption{An example of team traveling for four team tournament}
\label{tab_tseq}
$A=$
\begin{tabular}{ |c|| c c c| c c c|  }
\hline 
Team No.& $w_1$ & $w_2$ & $w_3$ & $w_4$ & $w_5$ & $w_6$ \\ \hline \hline
 $1$ & 0 & 0 & 0 & 1 & 1 & 1 \\
 $2$ & 1 & 0 & 0 & 0 & 1 & 1 \\
 $3$ & 0 & 1 & 1 & 1 & 0 & 0 \\
 $4$ & 1 & 1 & 1 & 0 & 0 & 0 \\
\hline
\end{tabular}
\end{table}

It is clear to see that any pair of teams have different traveling sequences. Next, the team scheduling also consists of $n$ scheduling sequences. The scheduling sequence in table $B$ of team $i$  is a sequence of number from 1 to $n$ except number $i$. The element $B(i,j)$ represents team number playing against team $i$ on week $w_j$. For example, one possible scheduling sequence of four team tournament as shown in Table~\ref{s_seq} \\

\begin{table}[h]
\centering
\caption{One possible scheduling sequence of team no. $1$} \label{s_seq}
\begin{tabular}{ |c|| c c c| c c c|  }
\hline 
Team No.& $w_1$ & $w_2$ & $w_3$ & $w_4$ & $w_5$ & $w_6$ \\ \hline \hline
$1$ & 2 & 3 & 4 & 2 & 3 & 4  \\
\hline
\end{tabular}
\end{table}

\vspace{10mm}
From the Table~\ref{s_seq}, it is a scheduling sequence of team $1$ and there is no number 1 in team $1$'s scheduling sequence. In general, there is no number $i$ in team $i$'s scheduling sequence. After all scheduling sequences are obtained, the team scheduling can be formed. An example of team scheduling of four team tournament is shown in Table~\ref{tab_sseq}.

\begin{table}[h]
\centering
\caption{An example of team scheduling for four team tournament}
\label{tab_sseq}
$B=$
\begin{tabular}{ |c|| c c c| c c c|  }
\hline 
Team No.& $w_1$ & $w_2$ & $w_3$ & $w_4$ & $w_5$ & $w_6$ \\ \hline \hline
 $1$ & 2 & 3 & 4 & 2 & 3 & 4 \\
 $2$ & 1 & 4 & 3 & 1 & 4 & 3 \\
 $3$ & 4 & 1 & 2 & 4 & 1 & 2 \\
 $4$ & 3 & 2 & 1 & 3 & 2 & 1 \\
\hline
\end{tabular}
\end{table}

From Table~\ref{tab_sseq}, element in each $B(i,j)$ depends on the team traveling in Table~\ref{tab_tseq}. For example, $B(1,3)$ can be equal to 4 because $A(1,3)$ is not equal to $A(4,3)$. In general, $B(i,j)$ can be equal to $k$ if and only if $A(i,j)$ is not equal to $A(k,j)$, since for each game, one team plays a home game and another team plays an away game. \\

Moreover, the sport tournament problem is solved by using the genetic algorithm together with a swapping method. A swapping method is used to generate a new traveling sequence. This method swaps the home/away roles of all games in a traveling sequence. The Figure~\ref{swap} shows 
the traveling sequences before and after one swapping.
\clearpage
\begin{figure}[h]
\centering
before:
\begin{tabular}{ | c c c| c c c|  }
\hline
 1 & 0 & 0 & 0 & 1 & 1 \\
\hline
\end{tabular} \\
after:     \ \
\begin{tabular}{ | c c c| c c c|  }
\hline
 0 & 1 & 1 & 1 & 0 & 0 \\
\hline
\end{tabular}
\caption{Traveling sequences before and after one swapping} \label{swap}
\end{figure}

In addition, an individual in GA is found by generating $\frac{n}{2}$ traveling sequences first. Then another $\frac{n}{2}$ traveling sequences are generated by using a swapping method. An example of generating an individual in case six teams as shown below. Firstly, the first three traveling sequences are generated by randomly choose with the TTP's condition. An example of this step is shown in Table~\ref{gtab_6t1}.
\begin{table}[h]
\centering
\caption{Generating the first three traveling sequences for six team tournament}
\label{gtab_6t1}
$B$=
\begin{tabular}{ |c|| c c c c c| c c c c c|  }
\hline
Team No.& $w_1$ & $w_2$ & $w_3$ & $w_4$ & $w_5$ & $w_6$ & $w_7$ & $w_8$ & $w_9$ & $w_{10}$ \\ \hline \hline
 $1$ & 0 & 0 & 1 & 1 & 0 & 1 & 1 & 0 & 0 & 1 \\
 $2$ & 1 & 0 & 1 & 0 & 1 & 0 & 1 & 0 & 1 & 0 \\
 $3$ & 1 & 0 & 1 & 1 & 0 & 0 & 1 & 0 & 0 & 1 \\ \hline
 $4$ &  &  &  &  &  &  &  &  &  & \\
 $5$ &  &  &  &  &  &  &  &  &  & \\
 $6$ &  &  &  &  &  &  &  &  &  & \\
\hline
\end{tabular}
\end{table}

After that, the remaining traveling sequences of team $4$, $5$, and $6$ are generated by using a swapping method with the traveling sequences of team $1$, $2$, and $3$, respectively. An example is shown in Table~\ref{gtab_6t2}.
\begin{table}[h]
\centering
\caption{Generating the remaining three traveling sequences for six team tournament}
\label{gtab_6t2}
$B$=
\begin{tabular}{ |c|| c c c c c| c c c c c|  }
\hline
Team No.& $w_1$ & $w_2$ & $w_3$ & $w_4$ & $w_5$ & $w_6$ & $w_7$ & $w_8$ & $w_9$ & $w_{10}$ \\ \hline \hline
 $1$ & 0 & 0 & 1 & 1 & 0 & 1 & 1 & 0 & 0 & 1 \\
 $2$ & 1 & 0 & 1 & 0 & 1 & 0 & 1 & 0 & 1 & 0 \\
 $3$ & 1 & 0 & 1 & 1 & 0 & 0 & 1 & 0 & 0 & 1 \\ \hline
 $4$ & 1 & 1 & 0 & 0 & 1 & 0 & 0 & 1 & 1 & 0 \\
 $5$ & 0 & 1 & 0 & 1 & 0 & 1 & 0 & 1 & 0 & 1 \\
 $6$ & 0 & 1 & 0 & 0 & 1 & 1 & 0 & 1 & 1 & 0 \\
\hline
\end{tabular}
\end{table}

\subsection{Algorithm}
\label{Algo}

The algorithm in this work bases on the genetic algorithm that has mutation and crossover processes. This algorithm is used to find the optimization solution of the team traveling but the team scheduling is found by another algorithm called ``Schedule." By deleting row and column heads of table $A$, we obtain the $n\times (2n-1)$ matrix before applying the following algorithm.\\

In the first algorithm, the mutation is performed before crossover by randomly choosing a week from 1 to $n-1$ of a random team to change from either 1 to 0 or 0 to 1. After that a mirrored week will be changed by the same condition due to mirrored tournament property. Moreover, two weeks of swapping teams are also changed by swapping property. An example of mutation process is shown in Figure~\ref{mut_pro}. 
\clearpage
\begin{figure}[h]
	\centering
	\subfigure[Before mutation process\label{b_mut}]{\includegraphics[width=4.5cm]{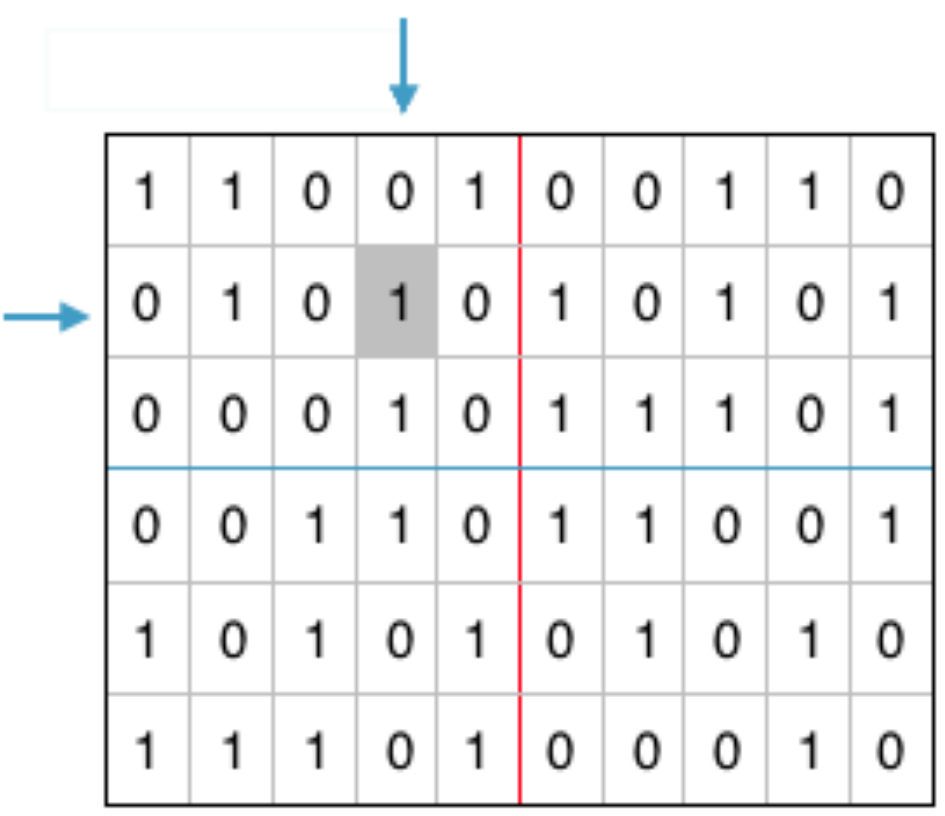}}
	\hfil
        \subfigure[After mutation process\label{logo3cm}]{\includegraphics[width=4cm]{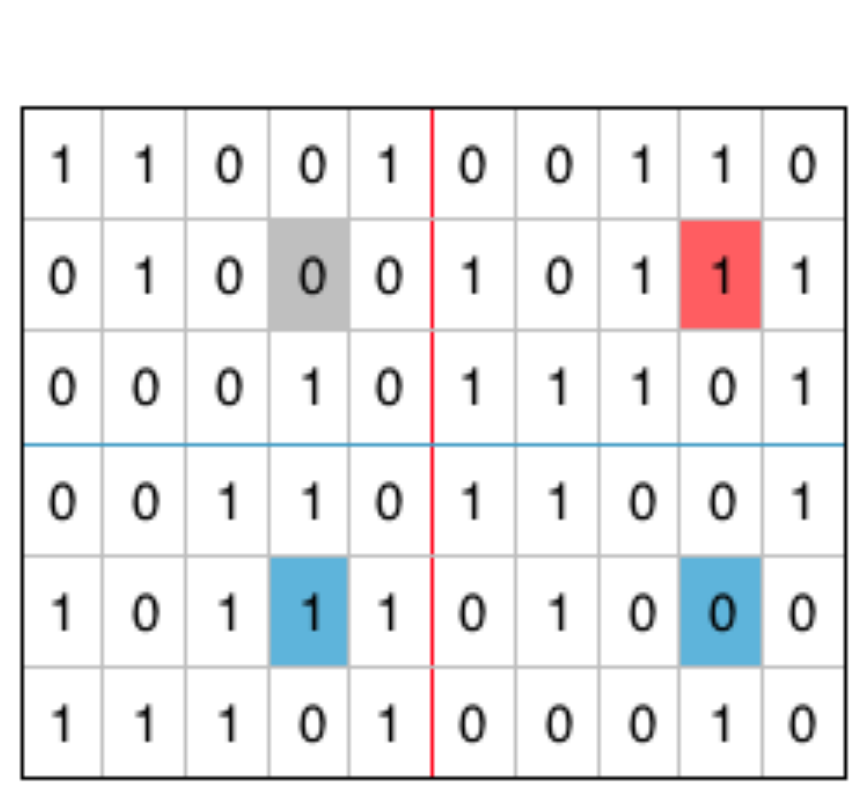}}
        \caption{The mutation process}
        \label{mut_pro}
\end{figure} 

Next, the crossover process is used to generate a new individual (team traveling) from two source individuals by the following steps:
\begin{itemize}
\item Find the ratio of each source individual to define a number of randomly chosen traveling sequences.
\item Randomly choose the first $\frac{n}{2}$ traveling sequences of a new individual depending on previous step.
\item After the first $\frac{n}{2}$ traveling sequences are obtained, the $\frac{n}{2}$ remaining traveling sequences are generated by using a swapping method.
\end{itemize}

Figure~\ref{n_cross} shows the crossover process of two individuals in case of six team tournament.

\begin{figure}[h]
\centering
\includegraphics[width=8cm]{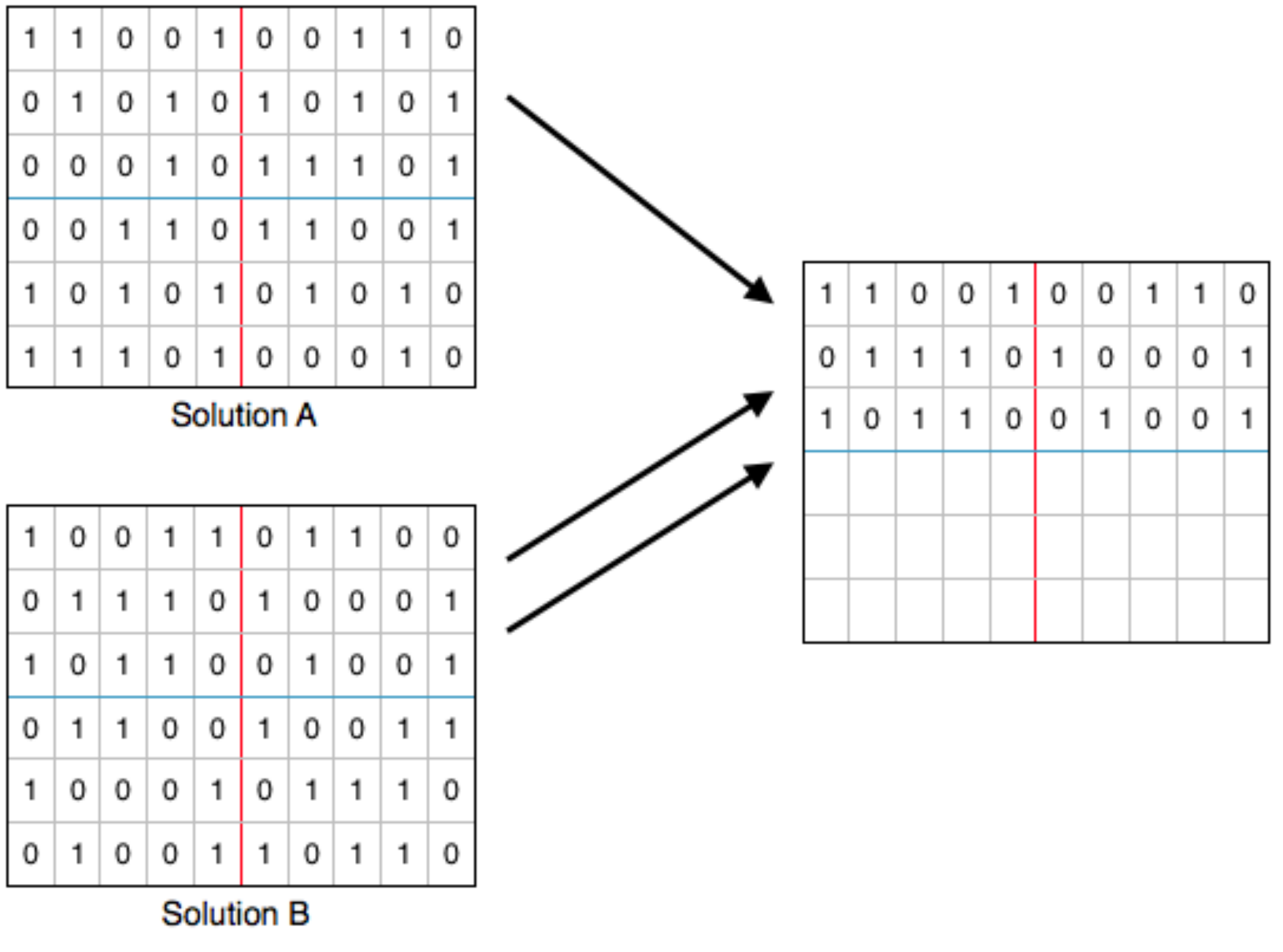}
\caption{A crossover process}
\label{n_cross}
\end{figure}

After the crossover process, the swapping process will be used to generate remaining teams of a new solution as shown in Figure~\ref{n_sol}.
\clearpage
\begin{figure}[h]
\centering
\includegraphics[width=4cm]{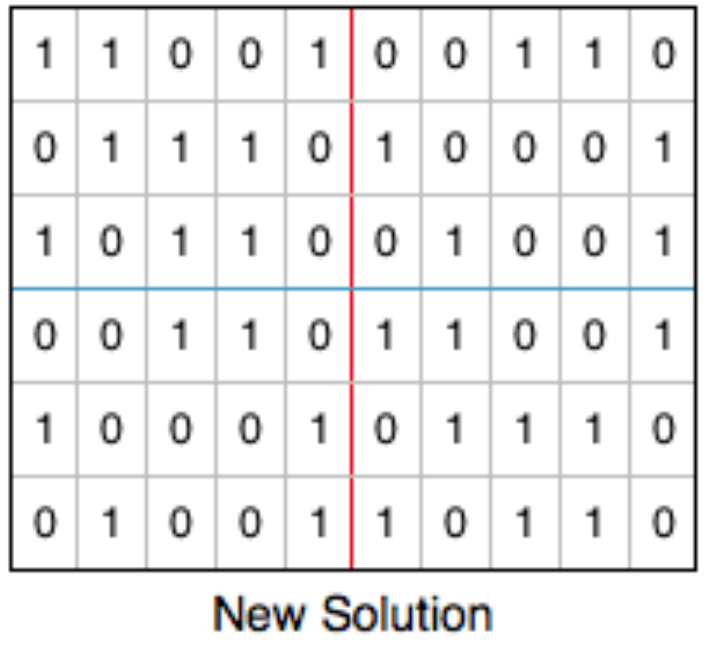}
\caption{A new solution}
\label{n_sol}
\end{figure}

After the optimized solution is obtained, using ``Schedule" algorithm to find the schedule of all teams. Figure~\ref{psub2} presents the pseudo code of ``Schedule" algorithm.

\begin{figure}[h]
\centering
\includegraphics[width=8cm]{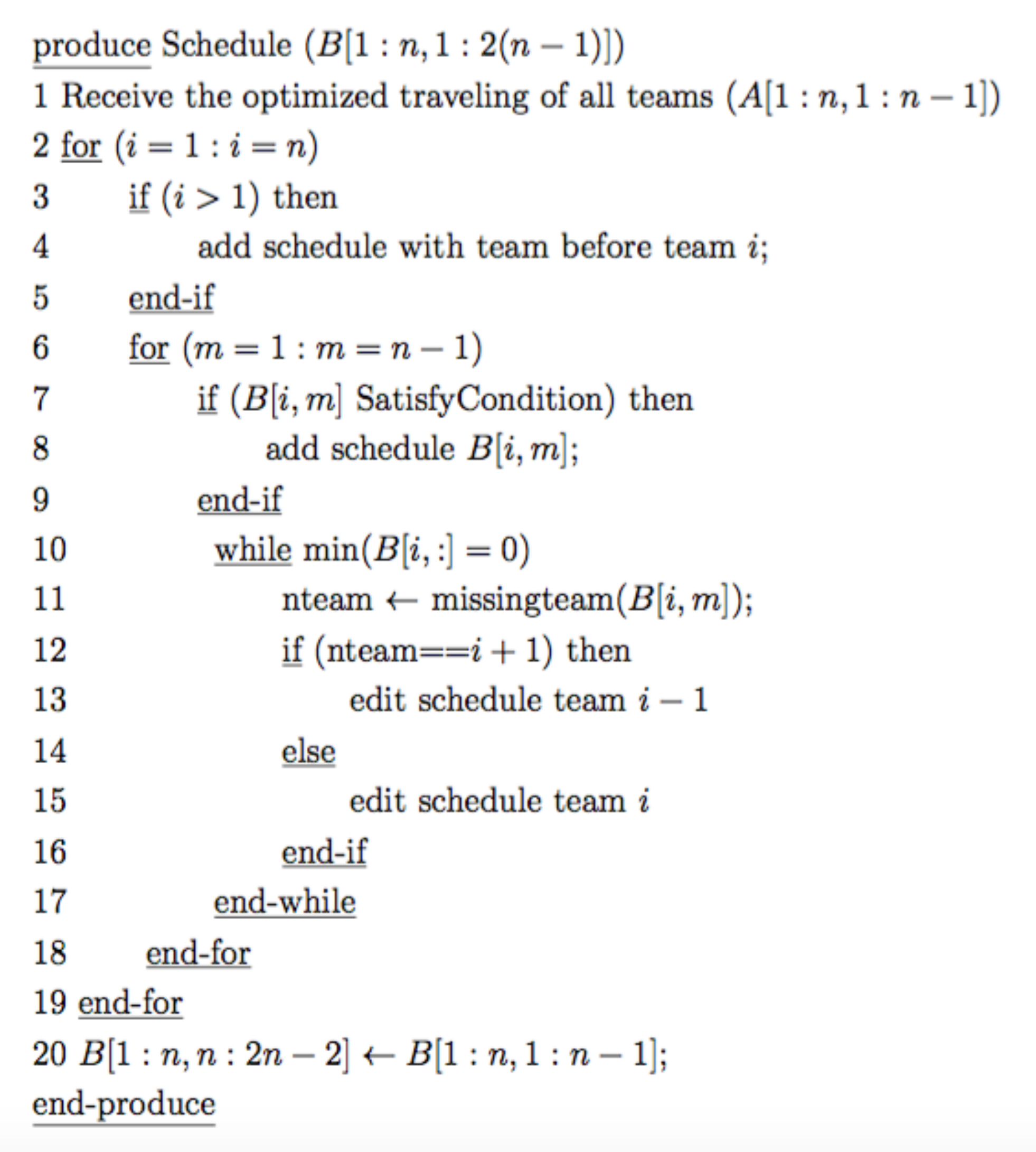}
\caption{Pseudo code of ``Schedule" algorithm}
\label{psub2}
\end{figure}

\newpage
\section{Experiments and Computational Results}
\label{m_result}

The algorithm in this work starts with four randomly populations. In each iteration, the two best solutions are kept and another two solutions are found by the mutation process is used with only $80\%$ probability. However, the crossover process are always used.\\

The table~\ref{tab_res} shows the results for any number of teams. The obtained results ($OR$) are compared with the known results ($KR$) and the lower bound ($LB$) of the mirrored traveling tournament problem \cite{ur06}. In addition, there are two relative gaps shown in the table~\ref{tab_res}. The first relative gap (gap$\# 1$) is the difference between the obtained results and the lower bound. The second gap (gap$\# 2$) is the difference between the obtained and the known results.

\begin{table}[h]
\centering
\caption{The computational Results}
\label{tab_res}
\begin{tabular}{ c c c c c c }
\hline
$n$	& $OR$ & $LB$ & $KR$ & Gap$\#1$ & Gap$\#2$ \\ 
	&	     &	         & 	     &	($OR-LB$)	 & ($OR-KR$)		       \\ \hline
 4	& 17 		& 17	 	& 17		& 0 & 0  \\
 6 	& 48 		& 48	 	& 48 		& 0 & 0  \\
 8	& 80 		& 80	 	& 80 		& 0 & 0  \\
 10	& 130 	& 130 	& 130	& 0 & 0  \\
 12	& 192 	& 192	& 192	& 0 & 0  \\
 14	& 253	& 252	& 256	& 1 & -3  \\
 16	& 348	& 342	& 342	& 6 & 6  \\
 18	& 432	& 432	& 434	& 0 & -2  \\
 20	& 521	& 520	& 526	& 1 & -5  \\
\hline
\end{tabular}
\end{table} 
\vspace{10mm}
The table~\ref{tab_res} shows that most obtained results approach to the lower bound in the literature and they are better than the known results.\\

\section{Conclusion}
\label{con}
In this work, we develop the genetic algorithm model by adding the swapping process in the crossover process to generate another traveling sequence that has the same or close to the number of traveling of source. The result also shows that the difference of number of traveling between any two teams is at most two.\\

In addition, the swapping process can reduce the running times because the algorithm will randomly choose $\frac{n}{2}$  traveling sequences from the two individuals after that the remaining traveling sequences will be generated by using the swapping process. The
obtained solution also satisfies the TTP's conditions. However, the algorithm without the swapping process must randomly choose $n$ traveling sequences from the two individuals and the obtained solution sometimes does not satisfy the TTP's conditions. It means that the algorithm must repeat the crossover process until the solution satisfies the TTP's conditions.\\

Moreover, a fair tournament defined by each team has the same number of traveling could be one of many constraints. This algorithm could make a fair tournament in case of some number of teams $n$. Nevertheless, a case of minimum number of traveling may not be the fair tournament.

\nocite{*}
\bibliographystyle{abbrvnat}
\bibliography{manuscript}
\label{sec:biblio}

\end{document}